\documentclass[sigconf,nonacm]{acmart}
\AtBeginDocument{%
  }

\setcopyright{rightsretained}
\copyrightyear{2025}
\acmYear{2025}
\acmConference{SIGGRAPH Posters '25}{August 10-14, 2025}{Vancouver, BC, Canada}
\acmBooktitle{Special Interest Group on Computer Graphics and Interactive Techniques Conference Posters (SIGGRAPH Posters '25), August 10-14, 2025}
\acmDOI{10.1145/3721250.3742965}

\newcommand{\acronym}{SEE-2-SOUND}

\newcommand{\humaneval}{43}

\usepackage{cleveref, float, xcolor, colortbl}

\usepackage{amsmath,amsfonts,bm}
\usepackage{mathtools}

\def\eqref#1{equation~\ref{#1}}

\def\1{\bm{1}}

\DeclareMathAlphabet{\mathsfit}{\encodingdefault}{\sfdefault}{m}{sl}
\SetMathAlphabet{\mathsfit}{bold}{\encodingdefault}{\sfdefault}{bx}{n}

\def\gC{{\mathcal{C}}}
\def\gD{{\mathcal{D}}}

\def\gM{{\mathcal{M}}}

\def\gP{{\mathcal{P}}}

\def\gR{{\mathcal{R}}}

\def\sR{{\mathbb{R}}}

\newcommand{\E}{\mathbb{E}}

\newcommand{\Eb}[2]{\E_{#1}\!\left[#2\right]}

\newcommand{\bA}{\mathbf{A}}
\newcommand{\bI}{\mathbf{I}}

\newcommand{\bc}{\mathbf{c}}

\newcommand{\bs}{\mathbf{s}}
\newcommand{\br}{\mathbf{r}}
\newcommand{\bx}{\mathbf{x}}

\newcommand{\bz}{\mathbf{z}}

\newcommand{\bepsilon}{{\boldsymbol{\epsilon}}}

\begin{document}

\title{\acronym: Zero-Shot Spatial Environment-to-Spatial Sound}


\author{Rishit Dagli}
\email{rishit@cs.toronto.edu}
\orcid{0000-0002-7622-3222}
\affiliation{%
  \institution{University of Toronto}
  \city{Toronto}
  \country{Canada}
}
\author{Shivesh Prakash}
\email{shivesh.prakash@mail.utoronto.ca}
\orcid{0009-0007-4120-0921}
\affiliation{%
  \institution{University of Toronto}
  \city{Toronto}
  \country{Canada}
}
\author{Robert Wu}
\email{rupert@cs.toronto.edu}
\orcid{0009-0005-6465-0154}
\affiliation{%
  \institution{University of Toronto}
  \city{Toronto}
  \country{Canada}
}
\author{Houman Khosravani}
\email{h.khosravani@utoronto.ca}
\orcid{0000-0002-4059-9420}
\affiliation{%
  \institution{University of Toronto}
  \city{Toronto}
  \country{Canada}
}

\renewcommand{\shortauthors}{Dagli et al.}

\begin{abstract}

Generating combined visual and auditory sensory experiences is critical for the consumption of immersive content. Recent advances in neural generative models have enabled the creation of high-resolution content across multiple modalities such as images, text, speech, and videos. Despite these successes, there remains a significant gap in the generation of high-quality spatial audio that complements generated visual content. Furthermore, current audio generation models excel in either generating natural audio or speech or music but fall short in integrating spatial audio cues necessary for immersive experiences. In this work, we introduce \acronym, a zero-shot approach that decomposes the task into (1) identifying visual regions of interest; (2) locating these elements in 3D space; (3) generating mono-audio for each; and (4) integrating them into spatial audio. Using our framework, we demonstrate compelling results for generating spatial audio for high-quality videos, images, and dynamic images from the internet, as well as media generated by learned approaches.
\end{abstract}




\begin{teaserfigure}
    \centering
    \includegraphics[width=\textwidth]{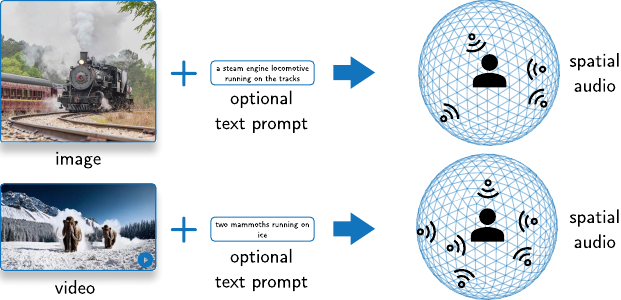}
    \caption{\acronym~enables spatial audio creation from visual content. Our method takes in an image or a video and generates spatial audio with a variable number of different sources. The audio could also be conditioned on a text prompt.}
    \Description{Showing an example input to and output from our framework.}
    \label{fig:teaser}
\end{teaserfigure}


\maketitle
\vspace{-0.3cm}
\section{Introduction}

Human visual and auditory systems inherently depend on synchronized spatial audio cues for an immersive experience and contextual understanding of video content. Modern generative models have been able to create compelling high-resolution digital content across various modalities. These models have been successful in generating images~\cite{NEURIPS2022_ec795aea, Rombach_2022_CVPR}, text~\cite{JMLR:v21:20-074, openai2024gpt4}, speech~\cite{45774, ren2021fastspeech}, audio~\cite{10530074, kreuk2023audiogen}, 3D scenes~\cite{Lin_2023_CVPR, poole2023dreamfusion}, 4D (dynamic 3D) scenes~\cite{singer2023textto4d, ling2024align}, and videos~\cite{singer2023makeavideo, videoworldsimulators2024}. Although generating audio from text~\cite{pmlr-v202-huang23i, copet2024simple} - and more recently, images~\cite{tang2023anytoany} - has become a reality, generating spatial audio is still a challenge. Unlike these earlier works, we are interested in solving the specific problem of generating high-quality spatial audio or surround sound to accompany images, animated images (GIFs), and videos. Such a model could be particularly useful to pair with any kind of video generation or image generation techniques as well as on real images and videos.

Audio generation is a long-standing challenge, often subdivided into generating speech, music, and sound effects. Most audio generation methods include inductive biases for each of these tasks, often due to the diversity between the generation of speech, music, and other types of audio~\cite{bresin2010expressive, agostinelli2023musiclm, 10.5555/3618408.3619294, tan2024naturalspeech}. There have also been newer approaches that develop unified representations~\cite{10530074} to generate complex audio as well. However, none of these currently emphasize the generation of rich spatial audio. The few related works that perform image-to-audio~\cite{tang2023anytoany, tang2023codi2} are closest to our work. However, these also face the same problem of not being able to generate rich spatial audio. Currently, there also exist a few learned approaches that generate spatial audio~\cite{NEURIPS2018_01161aaa, lin2021exploiting} but these are often based on having access to the captured audio; following which these approaches localized the audio in a viewing sphere to produce spatial audio for a captured video. However, none of these approaches aim to generate accompanying spatial audio for images, animated images (GIFs), and videos, which remains an open problem.

Towards this goal, we propose \acronym, a training-free method that approaches this specific problem by employing the foundational models, CoDi~\cite{tang2023anytoany}, DepthAnything~\cite{yang2024depth}, and Segment Anything~\cite{kirillov2023segment}. The pipeline works by explicitly factorizing this generative problem into separate problems: (1) identifying regions of interest in the input; (2) finding the 3D location on the viewing sphere for a visual element; (3) generating audio from the regions of interest, and (4) simulating these multiple mono-audio outputs to generate spatial audio. We then explicitly compose these component models to generate spatial audio conditioned on the visual input (e.g. images, animated images (GIFs), or videos) as shown in~\Cref{fig:teaser}. As our method is compositional and zero-shot, we hope to see \acronym\ widely used or adapted for content creation and enhancement as well as accessibility. 

By generating spatial audio that accurately corresponds to visual elements, we can significantly enhance the immersion of images and videos, creating more realistic and engaging user experiences with digital content.\acronym\ presents a step towards true complete video and image generation, by also incorporating non-verbal audio that accompanies visual generation models~\cite{singer2023makeavideo, videoworldsimulators2024, blattmann2023stable, Blattmann_2023_CVPR, luo2023videofusion, wang2024videocomposer}. Additionally, our pipeline can also be employed on real visual content.

Our contributions can be summarized as follows.
\begin{itemize}
    \item We present the challenging new task of generating spatial audio from images or videos. To the best of our knowledge, our work is the first of its kind that tackles this challenging problem.
    \item We propose \acronym, a novel compositional pipeline that builds on top of and uses off-the-shelf models for segmentation, depth estimation, and audio generation. Due to the nature of our approach, our method is zero-shot in nature and thus is very feasible to use.
    \item We evaluate \acronym\ both quantitatively and qualitatively through similarity comparisons and a user study, respectively. To evaluate this method quantitatively we also propose a novel evaluation method for this task.
\end{itemize}

We have open-sourced our code and models to promote the accessibility and reproducibility of our work. We also hope the extensibility of our work allows users to easily modify or replace individual modules within our approach. Our assets can be found at: \url{https://github.com/see2sound/see2sound}.

\section{Related works}
\label{sec:relatedworks}

The audio generation field has been gaining a lot ofmuch attention lately, primarily in models that create a diverse range of of audio content types. From speech synthesis to music generation and sound effects creation, past efforts have explored numerous aspects of auditory information synthesis. Learning algorithms have led to ever-improving techniques in the models for generating high-quality, diverse audio. This surge in interest has fueled research efforts pushing the boundaries of audio generation.

\subsection{Conditional Audio Generation}

Recent work in conditional audio generation has made significant progress using latent diffusion models~\cite{10.5555/3618408.3619294} such as AudioLDM (2)~\cite{10.5555/3618408.3619294, 10530074} and Make-An-Audio~\cite{pmlr-v202-huang23i}. Audiogen~\cite{kreuk2023audiogen} was one of the first to create a text-to-audio environment, treating audio generation as a conditional language modeling task. Building on this, AudioLDM-2~\cite{10530074} introduces a unified audio generation framework that addresses different objectives for speech, music, and sound effects, using the ``language of audio" (LoA) and a GPT-2~\cite{radford2019language} model for translation and latent diffusion~\cite{Rombach_2022_CVPR} for generation.

AudioLM~\cite{10158503} employs neural codecs~\cite{wang2023neural} for long-term consistency in audio generation, mapping input audio to discrete tokens and using a language modeling task. Meanwhile, DiffSound~\cite{yang2023diffsound}, using a text encoder, decoder, vector quantized variational autoencoder (VQVAE)~\cite{kingma2013auto}, and ``vocoder'', introduces a mask-based text generation strategy (MBTG) for generating text descriptions from audio labels to address the scarcity of audio-text paired data. NaturalSpeech~\cite{tan2024naturalspeech} integrates a memory component in variational autoencoders~\cite{kingma2013auto} to achieve human-level quality in text-to-speech systems.

Significant progress in text-to-music generation has been made by notable models such as MusicLM~\cite{agostinelli2023musiclm},\\ Noise2Music~\cite{huang2023noise2music}, MusicGen~\cite{copet2024simple} and MeLoDy~\cite{lam2023efficient}. MusicLM~\cite{agostinelli2023musiclm} uses contrastive pretraining to align music and language embeddings, improving performance without explicit text using a semantic modeling stage based on w2v-BERT~\cite{chung2021w2vbert}. MusicGen~\cite{copet2024simple} adopts a language modeling approach with melodic conditioning, enhancing control over output. MeLoDy~\cite{lam2023efficient} employs dual-path diffusion-based modeling for computational efficiency in music generation.

The image-to-audio and video-to-audio domains have also seen significant advancements. For instance, Im2Wav~\cite{sheffer2023i} generates audio from images or image sequences, using Transformers~\cite{NIPS2017_3f5ee243} to manipulate hierarchical discrete audio representation derived from VQ-VAE~\cite{oord2018neural}. Similarly, based on GANs~\cite{10.1145/3422622}: SpecVQGAN~\cite{iashin2021taming} enhances visually driven audio synthesis from videos, while RVQGAN~\cite{kumar2023highfidelity} achieves high-fidelity audio compression.~\cite{aytar2017see} introduces aligned representations that help in cross-modal retrieval and transferring between audio, visual, and text modalities. Furthermore, AV-NeRF~\cite{liang2024av} synthesizes new videos with spatial audio along different camera trajectories, and AV-RIR~\cite{ratnarajah2023av} estimates audio-visual room impulse responses from images and audio.

Multi-modal generative models of note include CoDi(-2)~\cite{Blattmann_2023_CVPR, tang2023codi2}. CoDi~\cite{Blattmann_2023_CVPR} generates multiple output modalities from various input combinations using a composable generation strategy. CoDi-2~\cite{tang2023codi2} extends this with enhanced modality-language alignment for encoding and generative tasks, improving zero-shot capabilities in multi-modal generation, reasoning, and interactive conversations.

Recent works like MAGNeT~\cite{ziv2024masked} introduce a masked generative sequence modeling method for direct audio token generation, achieving higher efficiency in text-to-music and text-to-audio tasks. DIFF-FOLEY~\cite{luo2024diff} improves video-to-audio synthesis quality, relevance, and synchronization using latent diffusion and contrastive audio-visual pretraining. Affusion~\cite{xue2024auffusion} enhances text-to-audio generation by leveraging text-to-image diffusion models, producing high-quality audio closely aligned with text prompts. These advancements significantly expand the possibilities of audio generation.

\subsection{Acoustic Matching}

Recent advancements in acoustic matching integrate visual and audio cues to enhance realism and immersion in generated content. AViTAR~\cite{Chen_2022_CVPR} performs visual-acoustic matching using a cross-modal transformer to integrate visual properties into audio and learning from in-the-wild videos. AdVerb~\cite{chowdhury2023adverb} uses a geometry-aware cross-modal transformer to enhance reverberation by leveraging visual cues, significantly outperforming traditional methods. SoundSpaces 2.0~\cite{chen2022soundspaces} provides geometry-based audio rendering for 3D environments, supporting tasks like audio-visual navigation and source localization with high fidelity and realism. 

\subsection{Construction of Surround Sound}

In 1881, Ader used a pair of microphones and telephone receivers to relay the sound signals at the Paris Opera~\cite{rumsey2012spatial}, a first-of-its-kind spatial (stereophonic) sound demonstration. In the past three decades, the development of digital signal processing~\cite{kirkeby1996local}, computer, and internet technologies have rapidly evolved spatial sound~\cite{xie2020spatial}. Together, they have made spatial audio commonplace now~\cite{zhang2017surround}.

In the context of virtual environments generated by game engines, spatial audio is crucial to provide an immersive experience~\cite{broderick2018importance}. Traditionally, sounds in virtual worlds are created by recording individual sound sources separately and then blending them according to the intended scene setup~\cite{murphy2011spatial} using various audio processing techniques~\cite{Gerzon1973PeriphonyWS, Dickins1999TowardsOS}. But this recording process is costly~\cite{zhang2017surround} and optimizing it involves generating spatial audio. While spatializing mono-audio is a popular area of research (discussed below), to the best of our knowledge, there have not been any approaches to generate spatial audio directly.

\subsection{Audio Spatialization}

Recent advancements in audio spatialization have focused on integrating audio and visual modalities to enhance realism. Sep-Stereo~\cite{zhou2020sepstereo} uses an associative pyramid structure to generate stereo output by combining audio and visual information. Similarly, LAVSS~\cite{ye2023lavss} introduces a location-guided audio-visual spatial audio separator, highlighting the importance of visual cues in audio generation.

Several studies have synchronized visual cues and encoders to produce natural sound for dynamic video scenarios~\cite{zhou2018visual}. Methods like those in~\cite{gan2019self, gao20192} use a U-Net to encode monaural input and decode binaural output with visual guidance.

Other approaches, such as~~\cite{gan2020foley, kreuk2023audiogen} generate audio samples conditioned on diverse inputs such as text, motion key points, and position information~\cite{garg2021geometry, parida2022beyond}, respectively. Techniques like~\cite{NEURIPS2018_01161aaa} utilize end-to-end trainable neural networks for sound source separation and localization in $360$-degree videos. Additionally, works like~\cite{li2018scene} synthesize directional impulse responses by exploiting the spatial and directional characteristics of room reverberation.

\begin{figure*}[ht!]
    \centering
    \includegraphics[width=\linewidth]{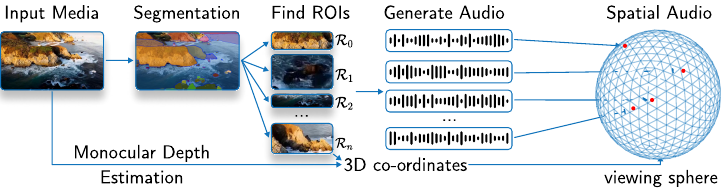}
    \caption{Overview of our pipeline. Given an input image we segment objects, estimate depth, generate per-object audio, then we place all of these audio according to the 3D co-ordinates on a viewing sphere producing spatial or surround sound.}
    \Description{Methods Figure.}
    \label{fig:methods_overview}
\end{figure*}

\subsection{Audio Visual Learning}

In recent years, there has been a growing interest in audio-visual learning, focusing on understanding the inherent connection between visual and auditory stimuli within vision research. Noteworthy works have delved into explaining the natural synchronization between visual and audio elements~\cite{arandjelovic2017look, aytar2016soundnet, owens2016ambient}. This approach finds application in diverse domains, including speech recognition utilizing both audio and visual cues~\cite{son2017lip, hu2016temporal, yu2020audio, zhou2019talking}, event localization through joint analysis of audio and visual information~\cite{tian2018audio, tian2020unified, wu2019dual}, and sound source localization~\cite{arandjelovic2018objects, hu2020discriminative, rouditchenko2019self, senocak2018learning, tian2018audio, zhao2018sound}. Additionally, there have been explorations in the use of audio-visual data for unsupervised representation learning, as demonstrated in~\citep{owens2018audio, owens2016ambient}. 

In contrast to these modalities, our focus diverges as we undertake a distinct task: the direct generation of surround sound spatial audio from images.

\section{Preliminary: Conditional LDM}

Latent Diffusion models (LDMs)~\cite{Rombach_2022_CVPR} project input images into a lower-dimensional latent space. Unlike traditional diffusion models that operate in a high-dimensional pixel space, LDMs perform forward and backward diffusion processes in a lower-dimensional learned latent space. For generating audio, we have a conditional LDM $\bepsilon_\theta$, which can be interpreted as a sequence of denoising autoencoders $\bepsilon_\theta(\bz_t, t, \ldots)$; $t \sim \mathcal{U}({1, \ldots, T})$, trained using a squared error loss (representing the denoising term in the ELBO) to denoise some variably noisy audio signals using the following objective,
\begin{equation}
\Eb{\mathcal{E}(\bx),\bc,\bepsilon \sim \mathcal{N}(0,I),t}{| \bepsilon - \bepsilon_{\theta}(\bz_t, t, \tau_\theta(\bc)) |^2_2},
\end{equation}

where $\bx_t$ is some noisy audio signal; $\bc$ is a conditioning vector; $\tau_\theta$ is an encoder that projects $\bc$ to an intermediate representation; $\mathcal{E}$ encodes $\bx_t$ into a latent representation $\bz_t$; and we jointly optimize $\bepsilon_\theta$ and $\tau_\theta$.

\section{Method}

Given a casually captured or generated image, animated image, or video, optionally with a text prompt that describes the content of the media, our objective is to generate spatial audio to accompany the media. We do not impose any qualitative restrictions on the way the media is captured. Our approach, \acronym\ is a compositional model that has three components (\Cref{fig:methods_overview}): (1) estimating positions for regions of interest (\S~\ref{sec:se}); (2) generating audio for regions of interest (\S~\ref{sec:ag}); and (3) generating spatial audio (\S~\ref{sec:ss}).

\subsection{Source Estimation}
\label{sec:se}

We have a visual input, for brevity let $\bI \in \sR^{H\times W\times C}$, represent a $(H, W)$-resolution and $C$ channel image. Here, we are interested in learning from $\bI$, a \emph{set} of 3D coordinates, $\bx = (x, y, z)$ representing the positions of regions of interest.

\paragraph{Identifying regions of interest.} We first focus on finding out regions of interest from the image, $\bI$. We do this by finding a segmentation mask $\gM\in\sR^{H\times W}$, using a pre-trained Segment Anything model~\cite{kirillov2023segment}. From the segmentation mask $\gM$, we extract the contours of each region of interest. Let $\gC = \{\bc_1, \bc_2, \ldots, \bc_n\}$ represent the set of contours, where each contour $\bc_k$ is a sequence of points outlining one region of interest. We then compute the smallest polygon $\gP_k$ that bounds each contour $\bc_k$. Formally, we solve the following optimization problem for each contour,
\begin{equation}
    \min_{\gP_k} \operatorname{Area}(\gP_k) \text{ , subject to, } \forall\bc_k \in\gP_k,
\end{equation}

where $\text{Area}(\gP_k)$ denotes the area of the polygon $\gP_k$, ensuring that the polygon fully encloses the contour $\bc_k$. This allows us to generate a set of minimal bounding polygons, $\gP = \{\gP_1, \gP_2, \ldots, \gP_k\}$.

We can also modify this to instead set each polygon to an axis-aligned bounding box (AABB) $\gR_k$ for each contour $\bc_k$. The AABB is defined by the minimum and maximum $x$ and $y$ coordinates of the points in $\bc_k$. Formally, for each contour $\bc_k$, we define $\gR_k$ as,
\begin{equation}
    \gR_k = \left[ \min(\bc_k^{(x)}), \max(\bc_k^{(x)}), \min(\bc_k^{(y)}), \max(\bc_k^{(y)}) \right].
\end{equation}

This allows us to generate a set of AABBs, $\gR = \{\gR_1, \gR_2, \ldots, \gR_k\}$.

\paragraph{Estimate monocular depth.} For the image $\bI$, we also estimate the monocular depth map $\gD\in\sR^{H\times W}$ using a pre-trained Depth Anything model~\cite{yang2024depth}. Now we can construct a set of image patches $\bI_\mathbf{p}$, for the regions of interest from either $\gP$ or $\gR$ to create a set of variably sized patches, and the 3D coordinates $(x, y, z)$, from the image coordinates and the monocular depth coordinates illustrated in~\Cref{fig:methods_overview}. We find that using AABBs usually works better for generating spatial audio.

\subsection{Audio Generation}
\label{sec:ag}

We are now interested in learning from the set of image patches $\bI_{\mathbf{p}}$ and an optional text prompt $\bc$, a \emph{set} of 4D representations, $(x, y, z, \bA^{T\times F})$ representing the positions and audio. We generate mono audio clips conditioned on each of the image patches in $\gR$ and the text conditioning $\bc$ using a pre-trained CoDi model~\cite{tang2023anytoany}. We thus have the 4D representation $(x, y, z, \bA^{T\times F})$ for each of the individual image patches in $\{\gR_1, \gR_2, \gR_3, \ldots\}$.

\subsection{Generating 5.1 Surround Sound Spatial Audio}
\label{sec:ss}

\begin{figure}[tb]
    \centering
    \includegraphics[width=\linewidth]{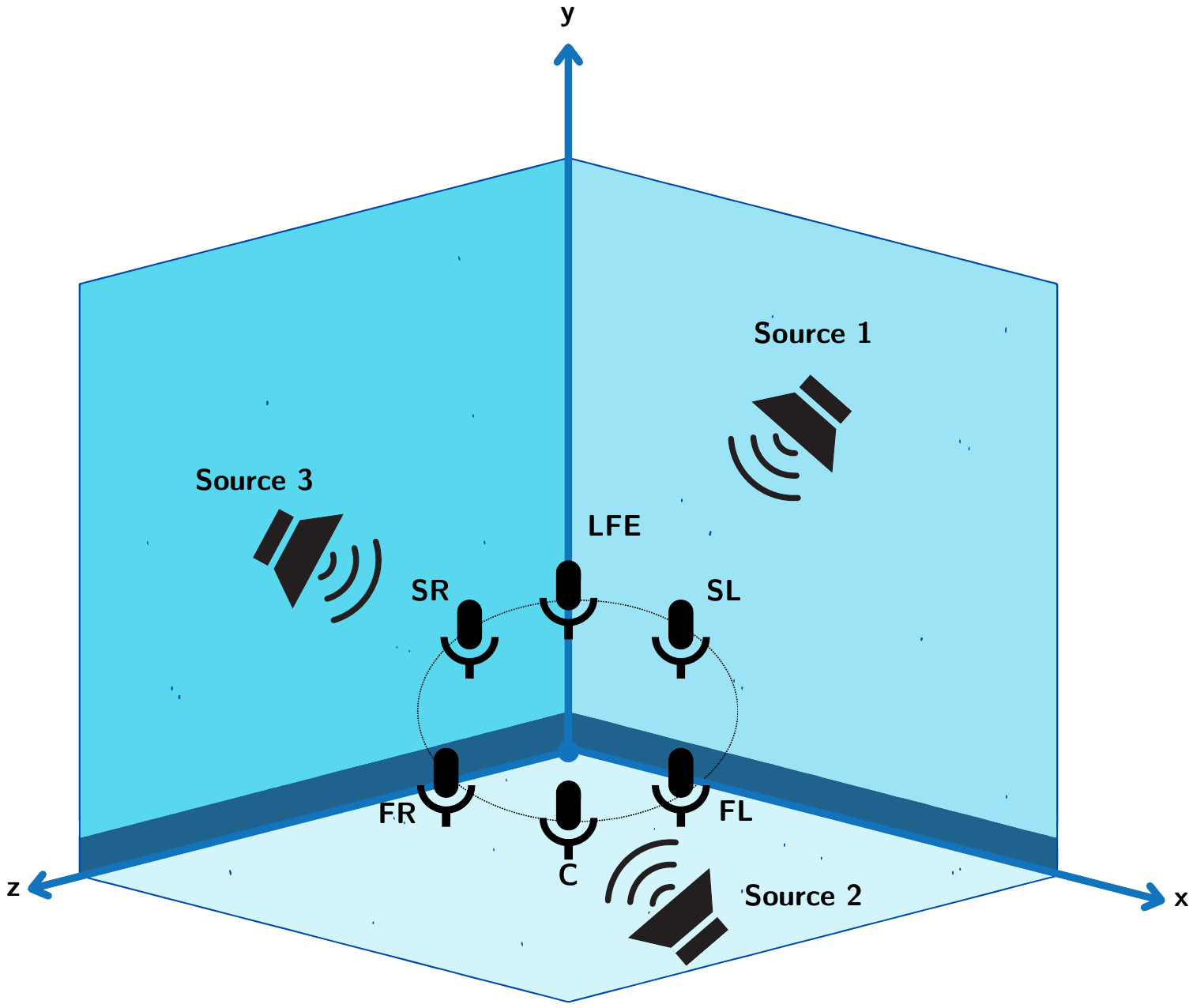}
    \caption{\textbf{Overview of the simulation.} We show a schematic of the virtual ``ShoeBox'' room utilized in our 5.1 surround sound simulation. The diagram presents the three-dimensional arrangement of multiple audio sources (labeled Source 1, Source 2, and Source 3) and the standardized microphone placement for the channels Front-Left (FL), Front-Right (FR), Center (C), Surround-Left (SL), Surround-Right (SR), and Low-Frequency Effects (LFE).}
    \Description{Representation of our room setup.}
    \label{fig:spatial_methods}
\end{figure}

The goal of surround sound spatial audio generation is to generate ambisonic channels $\phi_N(t)$ from different non-spatial audios and positions $(x, y, z, \bA^{T\times F})$. We utilize the pyroomacoustics library~\cite{scheibler2018pyroomacoustics} to simulate sound propagation between sources and receivers. While the library implements a ShoeBox room model, our implementation focuses solely on direct sound propagation without computing room impulse responses (RIRs) or reflections, making the room geometry irrelevant to the final output. We place the listener centrally at the ground level, serving as a reference point for audio projection and perception. For the source and microphone placements, we adhere to the ITU-R BS 775 configuration~\cite{series2010multichannel}, which is essential for aligning our simulations with the standard 5.1 Surround Sound setups, including designated positions for the Front-Left (FL), Front-Right (FR), Center (C), Surround-Left (SL), Surround-Right (SR) and Low-Frequency Effects (LFE) channels.

Our implementation focuses on direct sound propagation, where the signal at each receiver position is determined by the geometric relationship between sources and receivers. For a source at position $\bs$ and a receiver at position $\br$, the received signal is expressed as:
\begin{equation}
    \phi_{ch}(t) = \sum_{i} \alpha_{i} \bA_i^{T\times F}(t - \tau_{i}),
\end{equation}

where $\tau_{i} = \frac{\|\br - \bs_{i}\|}{c}$ represents the propagation delay from source $i$ to the receiver, $c$ is the speed of sound, and $\alpha_{i}$ represents the attenuation due to distance. This model effectively captures the spatial distribution of sound sources without considering environmental reflections, making it suitable for both indoor and outdoor scenes.

The synthesized output incorporates the interference patterns between multiple audio sources, which is a significant factor that differentiates our approach from scenarios where fewer sources are used and then combined. The resulting sound field exhibits unique characteristics that arise from the simultaneous interaction of all sources, creating spatial audio properties that are intrinsically linked to the arrangement $(x, y, z)$ and the number of sources in the complete setup.

\section{Experiments and Results}

To the best of our knowledge, no previous techniques address the problem of spatial sound generation from visual content, thus we propose a novel quantitative evaluation technique based on marginal scene guidance as well as perform other qualitative evaluations (\Cref{sec:avitar}) alongside human evaluation (\Cref{sec:humaneval}). We demonstrate qualitative examples in the supplementary.

\subsection{Quantitative Evaluation}
\label{sec:avitar}

\begin{table}[tb]
  \centering
  \caption{\textbf{Quantitative Evaluation.} Scores are computed between each method and its AViTAR~\cite{Chen_2022_CVPR}-enhanced version. Lower MFCC-DTW and higher ZCR, Chroma, and Spectral values denote greater similarity.}
  \label{tab:evaluation_results}
  \begin{tabular}{lcccc}
      \toprule
      \rowcolor{blue!15}(+ AViTAR) & DTW $\downarrow$ & ZCR $\uparrow$ & Chroma $\uparrow$ & Spect $\uparrow$\\
      \midrule
      CoDi~\cite{tang2023anytoany} & 0.0008 & 0.80 & 0.70 & 0.85\\
      \acronym                     & 0.00003 & 0.95 & 0.77 & 0.95\\
      \bottomrule
  \end{tabular}
\end{table}

\begin{table}[tb]
  \centering
  \caption{\textbf{Image–Audio CLIP Similarity.}}
  \label{tab:cs}
  \begin{tabular}{lc}
      \toprule
      \rowcolor{blue!15}Method & CLIP $\uparrow$ \\
      \midrule
      Wav2CLIP~\cite{wu2022wav2clip} & 0.0247\\
      Im2Wav~\cite{sheffer2023i} & 0.0108\\
      \acronym                       & 0.0259\\
      \bottomrule
  \end{tabular}
\end{table}

We present a new metric to conduct a quantitative evaluation of the scene guidance of generations produced by our model using various audio similarity metrics. We first sample images from Laion400M~\cite{NEURIPS2022_a1859deb} and the web. We then generate audio from these images using our method as well as CoDi~\cite{tang2023anytoany} which serves as a baseline. We then use the audio outputs and images to generate modified scene-guided audio generated using AViTAR~\cite{Chen_2022_CVPR}. The use of an unconventional model-based evaluation stems from the lack of existing evaluations or any strong baselines for image-to-spatial audio. We report the average audio similarity scores between multiple such generated audio in~\Cref{tab:evaluation_results}. We particularly note that \acronym\ consistently generates spatial audio that matches the scene's content often causing AViTAR~\cite{Chen_2022_CVPR} to only slightly altering the generated audios. We show that \acronym\ performs significantly better in handling sound directiosn than other baselines.

We compute CLIP-score~\cite{wu2022wav2clip} to quantify the semantic similarty of the generated audio with the input image in~\Cref{tab:cs}. We sample 1000 images from the COCO~\cite{lin2014microsoft} test set. While the CLIP score only measures the closeness of mono-audio with the image and our method spatial audio, we still demonstrate the CLIP score for \acronym. \acronym\ is comparable to other previous state-of-the-art methods on CLIP score.

\subsection{Human Evaluation}
\label{sec:humaneval}

\begin{table}[tb]
  \centering
  \caption{\textbf{Human Evaluation.}  $\chi^{2}$-tests with $p<0.05$ indicate significant deviation from a uniform response distribution (lower is better).}
  \label{tab:humaneval}
  \begin{tabular}{lc}
      \toprule
      \rowcolor{blue!15}Metric & $p$-value $\downarrow$\\
      \midrule
      \multicolumn{2}{l}{\textbf{Spatial-Audio Quality}}\\
      \quad Realism              & $4.174\times10^{-5}$\\
      \quad Immersion            & $1.036\times10^{-4}$\\
      \quad Accuracy             & $1.036\times10^{-4}$\\
      \quad Clarity              & $5.927\times10^{-6}$\\
      \quad Consistency          & $2.089\times10^{-6}$\\[2pt]
      \multicolumn{2}{l}{\textbf{Audio Identification}}\\
      \quad Overall localisation & $4.523\times10^{-4}$\\
      \quad Direction detection  & $2.880\times10^{-7}$\\
      \quad Distance estimation  & $3.103\times10^{-2}$\\[2pt]
      \textbf{Adversarial Identification} & $9.345\times10^{-2}$\\
      \textbf{Audio Matching}            & $7.431\times10^{-5}$\\
      \bottomrule
  \end{tabular}
\end{table}

Due to the absence of traditional metrics, we also use human evaluation to evaluate our approach. To promote future research in this field, we have open-sourced the evaluation protocol for human evolution. We conduct human evaluations by surveying $\humaneval$ human evaluators, for which we use randomly selected images and videos from the internet that are non-iconic in nature. We use this qualitative evaluation technique specifically due to the absence of any baseline methods for spatial generation. Additionally, the task of spatial audio generation is challenging since there are multiple possible ground truth spatial audios for a given image or video. This makes it difficult for us to use any kind of similarity metric.
More details on the tasks in our user study can be found in our supplementary. We evaluate the results for each of these individual subtasks and questions and conducted \(\chi^2\) tests to evaluate significance at the \(p<0.05\) level.

Our human evaluation task has the following key components.

\begin{description}
    \item[Spatial Audio Quality.] Human evaluators qualitatively rate the realism, immersion, and accuracy of two images or videos with spatial audio generated by our method. These aspects are measured using a semantic differential scale.
    
    \item[Direction Identification.] Evaluators estimate the sensation of direction and distance of an object given the generated spatial audio. This assesses the effectiveness of the spatial sound in conveying the sensation of sound originating from multiple directions.
    
    \item[Adversarial Identification.] Evaluators are given a random mix of generated spatial audio and mono audio. They are tasked with discriminating whether the given audio is indeed spatial audio, their accuracy is measured.
    
    \item[Audio Quality.] Evaluators are provided with three clips of spatial audio and three images or videos. They perform a matching task to pair the audio clips with the corresponding visual content, and their matching accuracy is measured.
\end{description}

We summarize our results in~\Cref{tab:humaneval}. We particularly note that human evlauation users indicated a statistically significant preference towards \acronym\ in terms of spatial audio quality, directions of sound, and audio quality.



\section{Discussion}
\label{sec:discussion}

Our work introduces a method for generating spatial sound from images and videos. We envision many avenues for potential impact based on our technique, including applications related to enhancing images and videos generated by learned methods, making real images interactive, human-computer interaction problems, or for accessibility.

\paragraph{Limitations.} Although \acronym\ produces compelling spatial audio clips, there are several limitations and avenues for future work. First, our framework may not detect all of the fine details in images and videos, nor does it produce audio for every fine detail. If we try to produce audio for many fine structures with this framework using upscaling or other zooming-in methods, we observe a lot of interference within the audios. 

Second, \acronym\ does not generate audios based on any motion cues but solely based on the image priors.  We could benefit from having separate appearance and motion backbones, which may potentially allow us to use any motion information to refine the generated audio. 

Third, currently, our method is not able to work in real-time on an Nvidia A100-80 GPU on which the experiments were conducted. However, since \acronym\ relies on composing off-the-shelf models, we could easily swap out the models we use for other models that solve the individual subproblems. We can easily incorporate advances in any of the subproblem areas within our framework to potentially get to real-time capabilities.

Lastly, inherent in the capabilities of generative models such as \acronym, which excel in producing synchronized and realistic outputs, is a significant concern regarding the potential generation and dissemination of deepfakes. The exploitation of this technology by malicious entities poses a notable risk even though we do not generate speech, as they could fabricate highly convincing counterfeit audio clips. We condemn any such application and hope to be able to work on safeguards for such approaches in the future.

\section{Conclusion}

Our work builds an audio-generation pipeline, \acronym, that can generate spatial audio from images, dynamic images, or videos. We achieve this by framing it as a compositional problem and propose a novel framework to use off-the-shelf methods for this challenging new task. To the best of our knowledge, our work is also the first to tackle the problem of performing spatial audio generation. Evaluating approaches for this novel task is very challenging thus apart from doing a user study we also propose a novel quantitative evaluation method based on measuring scene guidance from audios. We and our human evaluators observe compelling outputs for images and videos both from the internet and produced by generative models. Our method is also zero-shot in nature which makes our method significantly easy to use and extensible. To this extent, we also envision users being able to swap out the models we use for advances and other models in each of the individual sub-problems in our pipeline.

While not directly tested in this study, our work has significant implications for accessibility applications. The ability to generate spatial audio from visual content could potentially enhance the experience of visual media for individuals with visual impairments. Furthermore, this technology may have applications in medical settings, such as in neurology for patients recovering from stroke, and in broader accessibility and rehabilitation contexts. Future research could explore these potential applications to further understand and leverage the benefits of spatial audio generation in improving accessibility and quality of life for diverse user groups.

We hope \acronym\ and the future works we inspire lead to the generation of truly immersive digital content. We also look forward to seeing other research on the problem we describe in this work. Our codes, models and results are open-sourced to facilitate future research in this area.

\section*{Acknowledgements}

The authors thank Bhavya Bhatt of the University of Toronto for his help on surveying the existing literature. This research was enabled in part by the support provided by the Digital Research Alliance of Canada \footnote{\url{https://alliancecan.ca/}}. This research was supported in part with Cloud TPUs from Google's TPU Research Cloud (TRC)\footnote{\url{https://sites.research.google/trc/about/}}. The resources used to prepare this research were provided, in part, by the Province of Ontario, the Government of Canada through CIFAR, and companies sponsoring the Vector Institute \footnote{\url{https://vectorinstitute.ai/partnerships/current-partners/}}.

\bibliographystyle{ACM-Reference-Format}
\bibliography{sample-base}


\appendix

\clearpage
\newpage

\section{Details of our Method}
\label{sec:details}

We provide details of our approach with which we demonstrate all of our experiments and results. Our experiments ran on a single A100-80 GPU. Running our approach on an image takes $\sim12-15$ seconds (non-initial runs; initial runs can take up to $\sim3$ minutes) for images and $\sim5$ minutes on videos that are less than $20$ seconds like the ones we experiment with. Our method can be easily swapped for other off-the-shelf models, however we limit our experiments to these settings. Our human evaluation experiments were also performed without any text prompts to solely determine our framework's performance.

\paragraph{Source Estimation.} For source estimation we use a Segment Anything~\cite{kirillov2023segment} model. We use the ViT-H/16~\cite{dosovitskiy2021an} with $14\times 14$ windowed attention as our image encoder. Following standard approaches we rescale any inputs to $1024\times 1024$ by rescaling and padding. For the decoder, we use the same design and weights as used by Segment Anything~\cite{kirillov2023segment}. For monocular depth estimation, we use a Depth Anything~\cite{yang2024depth} model. We use ViT-L/14~\cite{dosovitskiy2021an} as our image encoder.

\paragraph{Audio generation.} For audio generation we use a CoDi~\cite{tang2023anytoany} model and use the Video LDM, Audio LDM, and Text LDM components. For inference, we sample the model through $500$ diffusion steps.

\paragraph{Generating 5.1 Surround Sound.}

We use a sampling rate of $16000$ Hz for audio processing. For simulating the acoustic environment, we utilize a ``ShoeBox" room model from Pyroomacoustics~\cite{scheibler2018pyroomacoustics}. The dimensions of the room are dynamically calculated based on the size of the image. Specifically, the dimensions width $w=$ image\_width, height $h=$ image\_height, and length $l= 0.5\cdot$ image\_width. In our virtual room, the microphones are placed at the following coordinates for each channel,
\begin{itemize}
    \item Front-Left (FL): [0, 0, 100]
    \item Front-Right (FR): [0, 0, -100]
    \item Center (C): [100, 0, 0]
    \item Low-Frequency Effects (LFE): [0, -100, 0]
    \item Surround-Left (SL): [-100, 0, 0]
    \item Surround-Right (SR): [0, 100, 0]
\end{itemize}

\section{User Study Protocol}
\label{sec:userstudy}

We demonstrate the user study that the participants were given. We however removed any YouTube video links we had as well as a consent form and institutional details here.

\subsection{Consent Form}
\label{sec:consent}

Your decision to participate is voluntary, and you are free to withdraw at any time without any consequences. You are not required to answer any questions that you do not feel comfortable answering.

Please ensure that you understand the following:
\begin{enumerate}
    \item The purpose of the survey is to evaluate AI for spatial audio generation from image inputs and optional text prompts for the purpose of improving the quality of the outputs from these systems.
    \item I will be asked a questionnaire consisting of comparisons between pairs of images, audios, and videos.
    \item The questionnaire takes approximately 15 minutes to complete, and will be conducted online.
    \item I may withdraw from the questionnaire at any time by exiting the questionnaire screen. If I choose to withdraw, all partial data collected will be discarded.
    \item The researchers do not foresee any risks or stresses on the participant beyond what one might experience in a typical day of work.
    \item The researchers also do not foresee any immediate benefits for the participant beyond what one might experience in a typical day of work.
    \item The results collected will help researchers develop systems for using AI models to generate spatial audio from image inputs and optional text prompts.
    \item I agree to have my responses documented and recorded digitally.
    \item All data collected from the questionnaire will be kept secure. Only the researchers involved in this study will have access to the information I provide. All data collected will be destroyed 5 years after publication of results.
    \item Data collected may be used in research journals, conferences, or other scholarly activities. In these cases, an anonymous subject identifier will be used and no identifying information will be provided to the audience.
    \item I am at least 18 years of age.
    \item I am free to ask questions about the process at any time. I can ask questions in person, or by contacting through email.
    \item If requested, I will receive a copy of this form for my records.
\end{enumerate}

The research study you are participating in may be reviewed for quality assurance to make sure that the required laws and guidelines are followed. If chosen, (a) representative(s) of the Human Research Ethics Program (HREP) may access study-related data and/or consent materials as part of the review. All information accessed by the HREP will be upheld to the same level of confidentiality that has been stated by the research team. If you have any questions about your rights as a participant, please contact the Ethics Review Office at (this email) or (this phone number).

\subsection{Part 1: Familiarization}

Before we begin, please note that the optimal testing condition involves 5.1 surround speakers. However, for this test, we ask participants to use headphones.

Watch this preselected video with correct spatial audio to familiarize yourself with the spatial audio experience.

\subsection{Part 2: Rating Spatial Audio with Images}

You will be exposed to images paired with corresponding spatial audio. Some of these may be generated by our system.

Rate the realism, immersion, accuracy, clarity, and consistency of the audio representation in relation to the visual content on a scale of 1 to 5, where 1 is not realistic/immersive/accurate/clear/consistent, and 5 is very realistic/immersive
/accurate/clear/consistent.

Here is how we have defined these terms:
\begin{itemize}
    \item \textbf{Realism:} How authentically the spatial audio mirrors real-world sound distribution.
    \item \textbf{Immersion:} The extent to which spatial audio engages and envelops the listener.
    \item \textbf{Accuracy:} The precision of alignment between spatial audio and visual content.
    \item \textbf{Clarity:} The degree to which individual sound elements are distinct and perceptible.
    \item \textbf{Consistency:} The uniformity of spatial audio representation across different images.
\end{itemize}

\textbf{Video 1} \\
Please rate the spatial audio in Video 1 on the following parameters, where 1 is not realistic/immersive/accurate/clear/consistent, and 5 is very realistic/immersive
/accurate/clear/consistent.

\textbf{Video 2} \\
Please rate the spatial audio in Video 2 on the following parameters, where 1 is not realistic/immersive/accurate/clear/consistent, and 5 is very realistic/immersive
/accurate/clear/consistent.

\subsection{Part 3: Direction and Distance Identification}
You will be exposed to videos paired with corresponding spatial audio. Some of these may be generated by our system.

Identify the direction and distance of sound sources within the audio landscape as depicted in the videos.

\textbf{Video 3}

\begin{enumerate}
    \item Where did you perceive the sound source for the trumpet to be located in the audio clip? (You can choose multiple options)
    \begin{itemize}
        \item Left
        \item Right
        \item Center
        \item Behind
        \item In front
        \item Above
        \item Below
    \end{itemize}

    \item How would you describe the distance of the sound source for the trumpet in the audio clip?
    \begin{itemize}
        \item Very Close
        \item Moderately Close
        \item Moderately Far
        \item Very Far
    \end{itemize}

    \item Where did you perceive the sound source for the drums to be located in the audio clip? (You can choose multiple options)
    \begin{itemize}
        \item Left
        \item Right
        \item Center
        \item Behind
        \item In front
        \item Above
        \item Below
    \end{itemize}

    \item How would you describe the distance of the sound source for the drums in the audio clip?
    \begin{itemize}
        \item Very Close
        \item Moderately Close
        \item Moderately Far
        \item Very Far
    \end{itemize}

    \item How accurately could you localize different sound sources within the audio landscape?
    \begin{itemize}
        \item Very Accurate
        \item Moderately Accurate
        \item Not Accurate
    \end{itemize}
\end{enumerate}

\subsection{Part 4: Real vs. Fake Identification}
You will be shown 2 clips, one of them has spatial audio generated by  and another one has mono audio of the same clips. You will be asked to identify which of them is "real" spatial audio and "fake" spatial audio.

\textbf{Video 4a}

\textbf{Video 4b}

Which of the clips is more likely to be real spatial audio?
\begin{itemize}
    \item Video 4a
    \item Video 4b
\end{itemize}

\subsection{Part 5: Matching Images to Spatial Audio}
You will be shown three images and corresponding spatial audio. Some of these may be generated by our system. You will be asked to match each spatial audio clip to its corresponding image.

Image 1 was shown.

\textbf{Spatial Audio 1}

Image 2 was shown.

\textbf{Spatial Audio 2}

Image 3 was shown.

\textbf{Spatial Audio 3}

Match each spatial audio clip to its corresponding image.

\section{Quantitative Evaluation Metrics}

The evaluation metrics we utilized are listed below:

\begin{enumerate}
    \item \textbf{MFCC-DTW Similarity}: Measures the similarity between the Mel-Frequency Cepstral Coefficients (MFCCs) using Dynamic Time Warping (DTW) distance.
    \item \textbf{Zero-Crossing Rate (ZCR) Similarity}: Measures the similarity between the zero-crossing rates of the audio files.
    \item \textbf{Rhythm Similarity}: Calculates the similarity between the rhythm patterns using the Pearson correlation coefficient.
    \item \textbf{Chroma Similarity}: Measures the similarity between the chroma features.
    \item \textbf{Spectral Contrast Similarity}: Compares the spectral contrast of the audio files.
    \item \textbf{Perceptual Similarity}: Quantifies the perceptual similarity using the Short-Time Objective Intelligibility (STOI) metric.
\end{enumerate}

We excluded the perceptual and rhythm similarity metrics for the following reasons:
\begin{itemize}
\item \textbf{Perceptual Similarity}: Comparing spatial audio to mono audio using perceptual similarity (such as the Short-Time Objective Intelligibility (STOI) metric) is not meaningful, as the perceptual qualities are inherently different.
\item \textbf{Rhythm Similarity}: Our framework does not generate music; hence, rhythm similarity is irrelevant in our context.
\end{itemize}

\section{Licenses}
\label{sec:licenses}
Here we list the licensing information for code bases, pre-trained models, and data used:

\begin{itemize}
    \item Segment Anything~\cite{kirillov2023segment} (code and weights): Apache License 2.0, \url{https://github.com/facebookresearch/segment-anything/tree/main}
    \item Depth Anything~\cite{yang2024depth} (code and weights): Apache License 2.0, \url{https://github.com/LiheYoung/Depth-Anything/tree/main}
    \item CoDi~\cite{tang2023anytoany} (code and weights): MIT License, \url{https://github.com/microsoft/i-Code/tree/main/i-Code-V3}
    \item Visual Acoustic Matching~\cite{Chen_2022_CVPR} (code and weights): Creative Common CC-BY-NC Licence, \url{https://github.com/facebookresearch/visual-acoustic-matching}
    \item Laion400M~\cite{NEURIPS2022_a1859deb}: Creative Common CC-BY 4.0 License, \url{https://laion.ai/blog/laion-5b/}
    \item Images from web: Creative Common CC-BY 4.0 License
\end{itemize}

Our code is made public under the Apache 2.0 License.

\end{document}